\documentclass[letterpaper]{article} 
\usepackage{aaai25}  
\usepackage{times}  
\usepackage{helvet}  
\usepackage{courier}  
\usepackage[hyphens]{url}  
\usepackage{graphicx} 
\def\UrlFont{\rm}  
\usepackage{natbib}  
\usepackage{caption} 

\usepackage[utf8]{inputenc} 
\usepackage[T1]{fontenc}    
\usepackage{url}            
\usepackage{booktabs}       
\usepackage{amsfonts}       
\usepackage{nicefrac}       
\usepackage{microtype}      
\usepackage{xcolor}         
\usepackage{enumitem}
\usepackage{graphicx}
\usepackage{multirow}
\usepackage{amsmath}
\usepackage[ruled,linesnumbered]{algorithm2e}
\usepackage{makecell}
\usepackage{caption}

\title{Collaborative Evolution: Multi-Round Learning Between Large and Small Language Models for Emergent Fake News Detection}

\author {
    Ziyi Zhou,
    Xiaoming Zhang\thanks{Xiaoming Zhang is the corresponding author.}, 
    Shenghan Tan,
    Litian Zhang,
    Chaozhuo Li
}
\affiliations {
    Beihang University,\\
    \{ziyizhou, yolixs, tsh21371459, litianzhang\}@buaa.edu.cn, lichaozhuo1991@gmail.com
}

\usepackage{bibentry}

\begin{document}

\maketitle

\begin{abstract}
The proliferation of fake news on social media platforms has exerted a substantial influence on society, leading to discernible impacts and deleterious consequences. 
Conventional deep learning methodologies employing small language models (SLMs) suffer from the necessity for extensive supervised training and the challenge of adapting to rapidly evolving circumstances. 
Large language models (LLMs), despite their robust zero-shot capabilities, have fallen short in effectively identifying fake news due to a lack of pertinent demonstrations and the dynamic nature of knowledge. 
In this paper, a novel framework Multi-Round Collaboration Detection (MRCD) is proposed to address these aforementioned limitations. 
The MRCD framework is capable of enjoying the merits from both  LLMs and SLMs by integrating their generalization abilities and specialized functionalities, respectively. 
Our approach features a two-stage retrieval module that selects relevant and up-to-date demonstrations and knowledge, enhancing in-context learning for better detection of emerging news events.  
We further design a multi-round learning framework to ensure more reliable detection results. 
Our framework MRCD achieves SOTA results on two real-world datasets Pheme and Twitter16, with accuracy improvements of 7.4\% and 12.8\% compared to using only SLMs, which effectively addresses the limitations of current models and improves the detection of emergent fake news detection.
\end{abstract}

\section{Introduction}
\label{introduction}

The proliferation of fake news on social networks has caused significant impact and harm to society~\cite{zhou2020survey}. In order to achieve automated detection of fake news, numerous deep learning-based approaches have been proposed~\cite{zhang2024reinforced,wu2023decor,hu2021compare}. The traditional methods primarily utilize small language models (SLMs) like BERT~\cite{devlin2018bert} to extract features from news content or propagation paths in social networks to accomplish the classification task, demonstrating promising performance across multiple fake news detection datasets.

\begin{figure*}[t]
\centering
\includegraphics[scale=1.0]{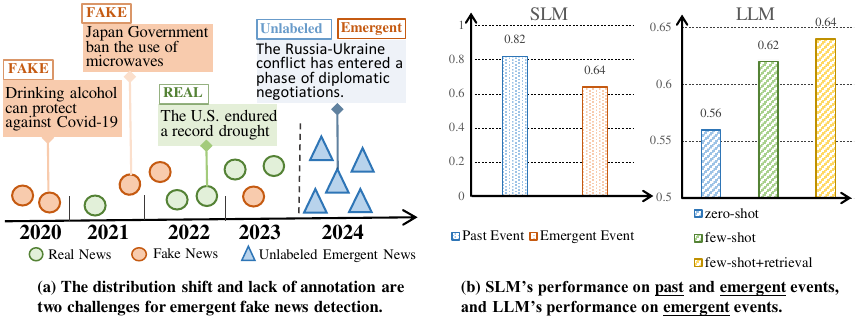}
    \setlength{\abovecaptionskip}{2mm}
    \setlength{\belowcaptionskip}{-4mm}
    \caption{(a) illustrates two major challenges for emergent fake news detection, the distribution shift and lack of annotations. (b) firstly demonstrates that SLMs perform well on testing past events after training, but its judgment capability significantly decreases on emergent events.  It then shows that LLM performs poorly on emergent events directly in zero-shot scenarios. However, its detection effectiveness improves after using  annotated data for few-shot learning and retrieval-augmented methods. The two experiments are conducted on Twitter16 dataset with RoBERTa as the SLM and Llama3-8B as the LLM.}
    \label{fig:f1}
\end{figure*}

Despite the satisfactory performance of current methods on specific datasets, these methods are mainly based on supervised training, whereas in reality, a large number of emergent news events occur frequently~\cite{olan2022fake}. Manual annotation of data is both time-consuming and costly, leading to insufficient data for supervised training of models~\cite{yin2024gamc,gangireddy2020unsupervised}. Moreover, the continuous evolution of news results in differences between the distribution of emergent events and manually annotated data as Figure~\ref{fig:f1}(a) illustrates. These factors collectively lead to traditional SLMs being unable to adapt well to emergent events, thus failing to demonstrate satisfactory performance in practical applications as Figure~\ref{fig:f1}(b) presented.

Large Language Models (LLMs) have shown remarkable abilities on various NLP applications due to their robust zero-shot capabilities~\cite{gruver2024large,kojima2022large,zhang2024interactive}. 
However, experiments in Figure~\ref{fig:f1}(b) have shown that LLMs do not exhibit strong detection capabilities for fake news in zero-shot and few-shot scenarios. This might be attributed to two main reasons: Firstly, while LLMs demonstrate strong task adaptation abilities, the potential of LLMs for fake news detection remains inactive due to the lack of suitable demonstrations in zero-shot environments. Secondly, news events are continually evolving while frozen-parameters LLMs lack external dynamic knowledge to assist in judging emerging events.


Due to the limitations of LLMs in fake news detection, recent researches~\cite{su2023adapting,wan2024dell,chen2023combating} have opted to utilize LLMs to provide additional assistance to SLMs for more accurate detection.
Dell~\cite{wan2024dell} utilizes LLMs as agents to form a social network by generating user comments and employ LLMs to extract additional information such as sentiment and knowledge to aid SLMs in fake news detection. ARG leverages LLMs to analyze news content and provide rationales to help SLMs in detection~\cite{hu2024bad}. Recent studies also leverage LLMs to extract more effective external knowledge to assist in assessing the authenticity of news~\cite{liu2024fakenewsgpt4}. 
Although the utilization of LLMs have shown improved detection capabilities, these methods still rely on a substantial amount of data for training the SLMs as they only use LLMs as an auxiliary to SLMs. 
Furthermore, they neglect the fact that LLMs possess strong learning and generalization capabilities for detecting fake news.

To address the dependency on data in current methods and the issue of current models' inability to accurately detect emergent events, we propose a novel \textbf{M}ulti-\textbf{R}ound \textbf{C}ollaboration \textbf{D}etection framework, dubbed \textbf{MRCD} that combines the generalization capability of LLMs with the specialized expertise of SLMs to achieve more accurate detection of the authenticity of emerging news events. Inspired by the success of in-context learning and retrieval-augmented generation across numerous downstream tasks~\cite{chen2023learning,sun2023pushing,lewis2020retrieval}, we also employ a novel two-stage retrieval module to select better demonstrations and acquire the latest knowledge relevant to the emergent news events.
Since there are no similar manually labeled news datasets available for emergent events, we use a news corpus and online search engines in the first-stage retrieval. This ensures we fetch the latest news articles closest to the news under detection. 
The extracted news content is then assigned with pseudo labels to serve as demonstrations for the LLM's in-context learning. The second-stage retrieval utilizes wikipedia as external knowledge corpus to retrieve latest knowledge relevant to the emergent events to provide the LLM more common sense information.
After performing in-context learning with the LLM and pre-trained SLM, we use both models to inference the unlabeled news articles and devise a data selection method to divide them into clean data pool and noisy data pool. Finally, we design a multi-round learning framework where the LLM and SLM collaborate to obtain a more generalized SLM and transform all data into samples with clean labels.
Our framework MRCD achieves SOTA results on two real-world datasets Pheme and Twitter16, with accuracy improvements of 7.4\% and 12.8\% compared to using only SLMs.

Our contributions are summarized as follows: 
\begin{itemize}[leftmargin=0.2cm, itemindent=0.2cm]
\item We introduce collaborative efforts between LLMs and SLMs for fake news detection and further devise a multi-round sample selection process to enhance detection accuracy in unsupervised emergent news events settings.
\item To enable the LLM to have a better semantic understanding of emergent news events and to access real-time knowledge, a two-stage retrieval module is proposed to select better demonstrations by utilizing online search engines and unlabeled news corpus while acquiring the latest and accurate knowledge from Wikipedia. 
\item Extensive experiments on real-world datasets demonstrate our method MRCD achieved SOTA performance and improves SLM's performance significantly by collaborative multi-round learning between the LLM and the SLM.
\end{itemize}

\begin{figure*}[t]
\centering
\includegraphics[scale=1.0]{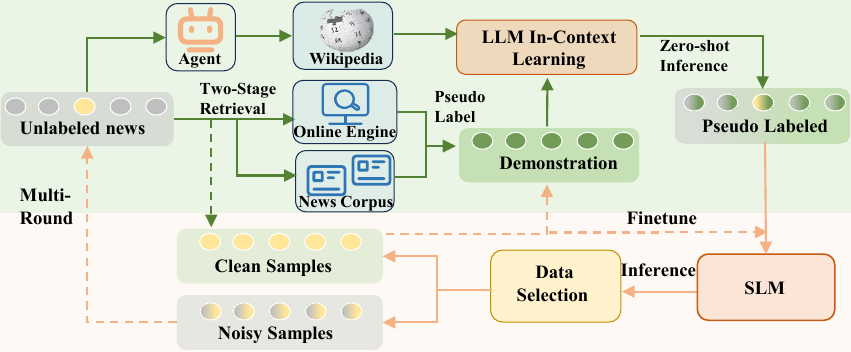}
    \caption{The architecture of MRCD. The $\rightarrow$ denotes the first round of learning while the $\dashrightarrow$ denotes the process for the subsequent rounds of learning.}
    \label{fig:model}
\end{figure*}

\section{Related Work}
\subsection{Fake News Detection}
Fake News Detection typically adopts a binary classification framework, distinguishing between real and fake news articles. Recent researches can be mainly divided into three categories: content-based methods, knowledge-augmented methods and propagation-based methods. Content-based methods primarily discern the authenticity of news by extracting distinct semantic information from real and fake news~\cite{khattar2019mvae,chen2022cross}. Knowledge-augmented methods utilize external knowledge corpus such as wikidata to provide common sense knowledge to aid model's detection~\cite{hu2021compare,zhang2024reinforced,sun2023inconsistent}. Propagation-based methods focuses on employing the propagation paths of news to identify misinformation~\cite{shu2020hierarchical,zhang2024mitigating,zhang2024early}.

Due to the remarkable capabilities of LLMs across various NLP tasks, recent research has emerged attempting to apply LLMs to fake news detection. Dell~\cite{wan2024dell} utilizes LLMs as agents to form a social network by generating user comments and employ LLMs to extract additional information such as sentiment and knowledge to aid SLMs in fake news detection. ARG leverages LLMs to analyze news content and provide rationales to help SLMs in detection~\cite{hu2024bad}. Recent studies also leverage LLMs to extract effective external knowledge to assist in assessing the authenticity of news~\cite{liu2024fakenewsgpt4}. Although these methods have shown promising performance, they have two main shortcomings.  Firstly, they still require a large amount of data to train the SLMs and lack good generalization ability for emerging news events~\cite{silva2021embracing,zhou2024finefake,zhang2023efficiently}. Secondly, they only utilize LLMs to provide additional knowledge to the small model, ignoring the large model's inherent ability for task reasoning and learning.

\subsection{In-Context Learning}
Recent research has demonstrated that LLMs can exhibit excellent performance in numerous zero-shot and few-shot downstream tasks through in-context learning~\cite{li2023finding,sun2023pushing,chen2023learning,ye2023compositional}. Among these researches, a considerable of them has focused on selecting better demonstrations~\cite{liu2022makes,wu2023self}. Unlike most studies that concentrate on utilizing algorithms like BM25 to select relevant data from annotated training datasets as demonstrations for test samples, Lyu and Min~\cite{min2022rethinking,lyu2023z} shows that in-context learning benefits mainly from the correct distribution of the inputs and the labels. Inspired by this, we choose to select demonstrations from both search engines and news corpus and assign pseudo labels for LLMs' in-context learning. This allows us to obtain examples closely related to the emergent news events under detection, avoiding significant distribution discrepancies between demonstrations and test data.

\subsection{Retrieval-Augmented LLMs}
Despite the outstanding performance of LLMs across various tasks, they still face challenges such as hallucination, outdated knowledge and high fine-tuning costs~\cite{gao2023retrieval,li2017ppne}. Retrieval-Augmented methods have shown promising effects in many knowledge-intensive tasks as they allow for continuous knowledge updates and integration of domain-specific information~\cite{lewis2020retrieval}, such as Question Answering~\cite{khattab2022demonstrate}, dialog generation~\cite{lin2023ra} and commonsense reasoning~\cite{cheng2023uprise}. To help LLM detect the authenticity of emergent news events, we employed a two-stage retrieval process to retrieve demonstrations and real-time knowledge for LLM.

\section{Problem Formulation}
This paper primarily aims to address the detection of emergent news events. Let $\mathcal{\varepsilon}$ denote a set of news events.  Each event $e$ is associated with a timestamp indicating its occurrence. We sort the events based on these timestamps, arranging them from earliest to latest : 
$\mathcal{\varepsilon} = {(e_1,t_1),(e_2,t_2),...,(e_n,t_n)}$. Then the events are divided into two past events and future events by their timestamps. The past events $\{X_e^s,Y_e^s\} = \{x_{e,i}^s,y_{e,i}^s\}_{i=1}^K$ are labeled posts and the data pertaining to future events $\{X_e^t\} = \{x_{e,i}^t\}_{i=K+1}^N$ lacks any form of annotation. The data of past events are used to initialize a SLM. The objective is for the LLM $\mathcal{L}$ and the initialized SLM $\mathcal{S}$ to collaborate effectively in detecting the authenticity of emergent news events.

\section{Methodology}
In this section, we introduce our proposed framework MRCD which investigates fostering collaboration between LLM and SLM. It incorporates a two-stage retrieval module to retrieve unlabeled news articles as demonstrations and utilizes knowledge from wikipedia to enhance the judgment capability of the LLM. 
The LLM then makes zero-shot predictions for the emergent news events under detection. Simultaneously, the pre-trained SLM directly infers pseudo-labels for the news. A data selection module is then designed to divide the data into clean and noisy subsets, with the noisy data undergoing another iteration for detection and the clean data using as demonstrations for the LLM and training samples for the SLM. The overview of MRCD is displayed in Figure ~\ref{fig:model}.

\subsection{Two-Stage Retrieval Module}
\subsubsection{Demonstration Retrieval}
Traditional demonstration selection often involves finding data related to the samples under detection from annotated training datasets for few-shot in-context learning~\cite{li2023finding,ye2023compositional,wu2023self,zhang2022deep}. However, due to significant distribution differences between continuously updated news events and pre-annotated training data, directly using annotated training data as examples may not enable the LLM to perform well in in-context learning. Recent research has also found that the success of in-context learning depends more on the semantic consistency between the test data and the demonstration data, rather than the correctness of the labels~\cite{liu2022makes,min2022rethinking,lyu2023z}. Therefore, we utilize online search engines\footnote{https://api.bing.microsoft.com/v7.0/news/search} $\mathcal{W}$ to retrieve the latest news, avoiding mismatches between data in static text corpus and emergent news events. To avoid the singularity of data selection from search engines, we additionally choose a news corpus~\cite{przybyla2020capturing,zhao2021gophormer} $\mathcal{C}$ to supplement and ensure diversity in the retrieved data.

Let $x$ denotes the news content, $N_w = \{w_1,w_2,...,w_n\}$ denotes the retrieved news from search engines $\mathcal{W}$ and $N_c = \{c_1,c_2,...c_m\}$ denotes retrieved news from news corpus $\mathcal{C}$. We utilize BM25 algorithm to extract the most semantically and structurally similar top-k data from $N_w$ and $N_c$ as demonstrations $N_k = \{d_1,d_2,...,d_k\}$ as Equation~\ref{bm25}: 
\begin{equation}
\label{bm25}
N_k = \text{BM25}_{\text{top-k}}(x , N_w \cup N_c) 
\end{equation}
Directly incorporating unlabeled news diminishes the LLM's ability to discern authentic news from fake, as the LLM relies heavily on labeled demonstrations to learn task-specific nuances~\cite{liu2022makes}. Additionally, simply assigning pseudo labels such as "real" or "fake" can result in copy effects, where the model mimics the surface characteristics of the labels rather than understanding the underlying truthfulness of the content~\cite{lyu2023z}. Inspired by~\cite{lyu2023z,li2018distribution}, we use semantic synonyms as pseudo labels. Each unlabeled instance $x_i$ is assigned a randomly selected label $\hat{y}_i \in \hat{\mathcal{Y}}$ to construct demonstrations $\mathcal{D}$:
\begin{equation}
\label{equation1}
\mathcal{D} = \{(x_1,\hat{y}_1),(x_2,\hat{y}_2),...,(x_k,\hat{y}_k) | x_i \in N_k, \hat{y}_i \in \hat{\mathcal{Y}}\}
\end{equation}
These synonyms are crafted to be semantically similar to 'real' and 'fake' but diverse enough to prevent direct replication of these terms by the LLM. By employing semantically rich alternatives, we ensure that the LLM engages more deeply with the content, activating its detection capabilities while preventing the copy effect.
The specific experiments on the usage and selection of synonyms as pseudo labels are discussed in detail in the Appendix A.4.

\subsubsection{Knowledge Retrieval}
Retrieval-Augmented methods have shown promising effects in many knowledge-intensive tasks as they allow for continuous knowledge updates and integration of domain-specific information~\cite{lewis2020retrieval,chang2024survey,zhang2022hierarchical}. To enable LLM to have a more detailed understanding of emergent news events and entities, we utilize wikipedia, which is continuously updated and widely used in knowledge-intensive tasks, as an external knowledge base to provide factual knowledge retrieval. 

We utilize a LLM as an agent to retrieve key entities $\{k_1,k_2,...,k_n\}$ from news content and then use the Wikipedia API\footnote{https://www.wikipedia.org/} to retrieve information about these key entities $\mathcal{K} = \{(k_1,i_1),(k_2,i_2),...,(k_n,i_n)\}$. The retrieved information with the news articles are provided to the LLM, enabling it to have the most up-to-date and accurate external knowledge assistance for understanding and judgment.

Once the demonstrations $\mathcal{D}$ and related knowledge $\mathcal{K}$ are determined, MRCD concatenates them with the content of the news article $x$ as input feed into the LLM. After in-context learning by demonstrations, LLM provides the predicted labels $\widehat{y_1}$ for $x$ supplemented with related knowledge:
\begin{equation}
    \widehat{y_1}=argmax_{\widehat{y_1} \in \mathcal{Y}}P(\widehat{y_1}| \mathcal{D},\mathcal{K},x)
\label{equation_llm}
\end{equation}
 The predicted labels $\widehat{y_1}$ from LLM and the news articles $x$ are passed to the SLM for further data filtering and labeling.

\begin{algorithm}[!t]
  \SetKwData{Left}{left}\SetKwData{This}{this}\SetKwData{Up}{up}
  \SetKwFunction{Union}{Union}\SetKwFunction{FindCompress}{FindCompress}
  \SetKwInOut{Input}{Input}\SetKwInOut{Output}{output}
  \Input{Emergent Events $\{X_e^t\} = \{x_{e,i}^t\}_{i=K+1}^N$, LLM $\mathcal{L}$, SLM $\mathcal{S}$ initialized with Past Events $\{X_e^s,Y_e^s\} = \{x_{e,i}^s,y_{e,i}^s\}_{i=1}^K$, round $=$ 1}
  \Output{Labeled Emergent Events $\{X_e^t\} = \{x_{e,i}^t,y_{e,i}^t\}_{i=K+1}^N$}
  \tcc{Multi-round Learning}\
  \If{round == 1}{
    Obtain Demonstrations $\mathcal{D}$ by BM25 and assign pseudo labels from $(\mathcal{W},\mathcal{C})$\;
    Knowledge-Retrieval Module to obtain $\mathcal{K}$\;
    Inference by $\mathcal{L}$ and $\mathcal{S}$ to obtain $(\widehat{y_1},\widehat{y_2})$\;
    $ \widehat{y_1}=argmax_{\widehat{y_1} \in \mathcal{Y}}P(\widehat{y_1}| \mathcal{D},\mathcal{K},x)  $\; $\widehat{y_2}=argmax_{\widehat{y_2} \in \mathcal{Y}}P(\widehat{y_2}|x)$
    \;
    Selection module to obtain  $D_{clean}$ and $D_{noisy}$\;
    round = round + 1\;
  }
  \For{round $\leq$ $\mathcal{N}$}{
    Retrieve  $\mathcal{D}$ for $\mathcal{L}$ and fine-tune $\mathcal{S}$ by $D_{clean}$\;
     $D_{noisy}$ inference by $\mathcal{L}$ and $\mathcal{S}$ to obtain $(\widehat{y_1},\widehat{y_2})$\;
     Selection module to update $D_{clean}$ and $D_{noisy}$\;
     round = round + 1\;
  }
  \If{$D_{noisy} \neq \emptyset$}{
    $D_{noisy}$ inference by $\mathcal{S}$ to obtain $\widehat{y_2}$\;
    Update $D_{clean}$ as final labels\;
  }
\caption{Pseudo-code for \textbf{MRCD}}
\label{alg1}
\end{algorithm}

\subsection{Data Selection Module}
As the news articles with pseudo labels $(x,\widehat{y_1})$ provided by the LLM are passed to the initialized SLM, the SLM directly infers pseudo labels $\widehat{y_2}$ for the news article $x$. In semi-supervised learning, using pseudo labels with high confidence as true labels for further training is a widely used approach~\cite{rizve2020defense,li2021adsgnn}. Inspired by this, we design a filtering mechanism that leverages the pseudo labels of both LLM and SLM to filter cleaner labels simultaneously.
All unlabeled news $\{(x,\widehat{y_1},\widehat{y_2}), x \in \mathcal{X})\}$ are divided into clean data pool $D_{clean}$ and noisy data pool $D_{noisy}$ by Equation \ref{equation2}, where $p(\widehat{y_2})$ denotes the output probability by SLM and $\omega$ denotes the confidence thresholds:
\begin{equation}
\begin{aligned}
D_{clean} = \{(x_i,y_i) \ &| \  \widehat{y_1} = \widehat{y_2} \ and \  p(\widehat{y_2}) \geq \omega\} \\
D_{noisy} = \{(x_i) \ &|\  \widehat{y_1} \neq \widehat{y_2} \ or \  p(\widehat{y_2}) < \omega\}
\end{aligned}
\label{equation2}
\end{equation}

\begin{table*}[t]
  \centering
    \resizebox{0.9\linewidth}{!}{
    \begin{tabular}{c|l|rrrr|rrrr}
    \toprule 
    \multirow{2}[4]{*}{Category} & \multicolumn{1}{c|}{\multirow{2}[4]{*}{Method}} & \multicolumn{4}{c|}{Pheme} & \multicolumn{4}{c}{Twitter16}  \\
\cmidrule{3-10}    \multicolumn{1}{c|}{}& \multicolumn{1}{c|}{} & \multicolumn{1}{c}{Acc} & \multicolumn{1}{c}{Pre} & \multicolumn{1}{c}{Rec} & \multicolumn{1}{c|}{F1} & \multicolumn{1}{c}{Acc} & \multicolumn{1}{c}{Pre} & \multicolumn{1}{c}{Rec} & \multicolumn{1}{c}{F1}  \\
    \midrule
    \multirow{4}[2]{*}{SLM} & RoBERTa  & 0.714      & \underline{0.777}      & 0.695      & 0.734      &0.644       &0.649       &0.641       &0.645         \\
          & EANN  & 0.744      & 0.738      & 0.745      & 0.741      & 0.641      & 0.621      & 0.741      & 0.676      \\
          & M$^{3}$FEND &  0.746     & 0.747      & 0.746      & 0.746      & 0.642      & 0.608      & \textbf{0.816}      & 0.697       \\
          & FTT &  0.754     & 0.748      & 0.764      & 0.756      & 0.651      & 0.649      & 0.720      & 0.683       \\
    \midrule
    \multirow{3}[2]{*}{\makecell[l]{LLM \\ (zero-shot)}} & Llama2-7B &0.505       & 0.498      & 0.697      & 0.581      &0.496       &0.501       &0.494        &0.498     \\
    & Llama3-8B & 0.535   & 0.518     & 0.770      & 0.620      &  0.562     & 0.574      &  0.531     &  0.552       \\
          & GPT3.5 &0.503       & 0.500      &  0.714     &0.586       & 0.583      &0.585       & 0.571      & 0.578      \\
    \midrule
    \multirow{3}[2]{*}{\makecell[l]{LLM \\ (few-shot)}}
          & Llama2-7B & 0.528      &0.511       &0.894       &0.650       &0.590       &0.598       &0.584       & 0.591     \\
          & Llama3-8B & 0.549      &0.524       & \textbf{0.961}      & 0.679      & 0.622      &0.607       & 0.717      &   0.658    \\
          & GPT3.5 &0.520       & 0.507      & 0.850      &0.635       &0.621       & 0.609      & 0.705      &0.653     \\
    \midrule
    LLM+SLM & ARG  & 0.743      & 0.741      & 0.779      & 0.760      & 0.705      &  0.698     & 0.710      &0.704         \\
    \midrule
    \multirow{4}[2]{*}{MRCD} & Llama2+RoBERTa &0.772       &0.765       &0.775       &0.770       &0.732       &0.717       & 0.619      & 0.664   \\
    & GPT3.5+RoBERTa  &0.781       & 0.735      &0.821       &0.778      &0.768     &0.752      & 0.734      & 0.743       \\
          & Llama3+RoBERTa
          &\underline{0.788}       & 0.700      & \underline{0.900}      &\underline{0.786}  &\underline{0.772}       &\underline{0.765}       &0.775    &\underline{0.770}      \\
          & Llama3+FTT
          &\textbf{0.814}       & \textbf{0.788}      & 0.841      &\textbf{0.814}  &\textbf{0.794}       &\textbf{0.768}       &\underline{0.782}       &\textbf{0.774}      \\
    \midrule
    \multirow{2}[2]{*}{Improvements}
    & \textit{Impr. RoBERTa} & +\textit{7.4\%}      & /      &  +\textit{20.5\%}     & +\textit{5.2\%}      & +\textit{12.8\%}     &+\textit{11.6\%}       & +\textit{13.4\%}     & +\textit{12.5\%}       \\
    & \textit{Impr. FTT} & +\textit{6.0\%}      & +\textit{4.0\%}      &  +\textit{7.7\%}     & +\textit{5.8\%}      & +\textit{14.3\%}     &+\textit{11.9\%}       & +\textit{13.2\%}     & +\textit{12.1\%}       \\
    \bottomrule
    \end{tabular}%
    }
    \caption{Performance of baselines and MRCD. Best results are in \textbf{bold} and second best results are \underline{underlined}. }
  \vspace{-0.4cm}
  \label{main_new}%
\end{table*}%

\subsection{Multi-Round Learning}
The initial classification of datasets into $D_{noisy}$ and $D_{clean}$ is only the beginning. Following this initial differentiation, a multi-round learning strategy is applied to further refine and enhance the quality of classifications, facilitating the dynamic adjustment of both the LLM and the SLM to capture emergent news efficiently.

Starting from the second round, these data in $D_{noisy}$ is re-evaluated as unlabeled data and also undergoes the process of two-stage retrieval and data selection. 
Distinct from the initial round where demonstrations are primarily retrieved from external sources, in subsequent rounds, demonstrations $\mathcal{D}^{'}$ are drawn from the previously established $D_{clean}$ utilizing the BM25 algorithm:
\begin{equation}
\begin{aligned}
\label{bm25_multiround}
N_k^{'} = \text{BM25}&_{\text{top-k}}(x , D_{clean}) \\
\mathcal{D}^{'} = \{(x_1,\hat{y}_1),(x_2,\hat{y}_2)&,...,(x_k,\hat{y}_k) | x_i \in N_k^{'} \}
\end{aligned}
\end{equation}

In addition, the SLM will also use the data with clean labels for fine-tuning, further enhancing its ability to detect emergent news events. Apart from the demonstration retrieval module and SLM's fine-tuning, the rest modules of MRCD remain the same as in the first round. 
Through iterative rounds, we employ an iterative re-labeling strategy where each piece of data in $D_{noisy}$ is reevaluated, and those meeting a specific confidence criterion are transferred to $D_{clean}$:
\begin{equation}
    \label{transfer}
    D_{clean}^{new} = \{ (x_i,y_i) | x_i \in D_{noisy} \ and \ p(\widehat{y_2}) > \omega \} \cup D_{clean}
\end{equation}
The newly validated instances in $D_{clean}^{new}$ are then used as both training data for the SLM and demonstrations for the LLM in the upcoming iterations.
When reaching the threshold $\mathcal{N}_{th}$ round, the remaining samples in $D_{noisy}$ are directly predicted by the SLM $\mathcal{S}$ as the final judgment labels, thus completing the iterative learning cycle and solidifying the dataset classification.
The overall pipeline is depicted in Algorithm ~\ref{alg1}, providing a detailed visualization of the multi-round learning.

\section{Experiments}
\subsection{Datasets}
To fairly evaluate the performance of the proposed model MRCD, we conduct experiments on datasets collected from real-world social media, namely Twitter16~\cite{boididou2018detection} and Pheme~\cite{zubiaga2017exploiting}. Twitter16 and Pheme directly crawled news articles from trending events on Twitter, with each piece of data labeled with the associated event. Therefore, in Pheme, we use the latest occurring events germanwings-crash as the test set. For Twitter16, we directly use the events in the test set for evaluation.
The detailed information about Twitter and Pheme are in Appendix A.2.

\begin{table*}[t]
  \centering
  \renewcommand{\arraystretch}{0.75}
    \begin{tabular}{c|c|rrrr|rr}
    \toprule
    Round & Model & \multicolumn{1}{c}{Accuracy} & \multicolumn{1}{c}{Recall} & \multicolumn{1}{c}{Precision} & \multicolumn{1}{c|}{F1-Score} & \multicolumn{1}{c}{Clean Pool} & \multicolumn{1}{c}{Noisy Pool} \\
     \midrule
    \multirow{2}[2]{*}{Round 0} & LLM(zero-shot)   & 0.562      & 0.574      &0.531       &0.552       & 0      &0  \\
          & LLM(few-shot)   & 0.622      & 0.607      &0.717       &0.658       & 0      &0  \\
    \midrule
    \multirow{2}[2]{*}{Round 1} & SLM   & 0.644      & 0.650      &0.641       &0.645       & 637      &789  \\
          & LLM   & 0.604      & 0.588      &0.722       &0.650       & 637      &789  \\
    \midrule
    \multirow{2}[2]{*}{Round 2} & SLM   & 0.678      & 0.662      &0.704       & 0.682    & 728      &698  \\
          & LLM   &0.674       &0.639       & 0.704      &0.678       &728       &698  \\
    \midrule
    \multirow{2}[2]{*}{Round 3} & SLM   &0.772       &0.765       & 0.775      & 0.770      &763       &663  \\
          & LLM   &0.665       &0.661       &0.739       & 0.689      &763       &663  \\
    \midrule
    \multirow{2}[2]{*}{{Round 4}} & SLM   &0.743       &0.713       & 0.791      & 0.751      &798       &628  \\
          & LLM   &0.663       &0.658       &0.692       & 0.674      &798       &628  \\
    \midrule
    \multirow{2}[2]{*}{{Round 5}} & SLM   &0.711    &0.703       & 0.803      & 0.749      &813       &613  \\
          & LLM   &0.667       &0.653       &0.695       & 0.671      &813       &613  \\
    \bottomrule
    \end{tabular}%
  \caption{Analysis on multi-round learning. 
  }
  \label{tab:multi_round}%
  \vspace{-4mm}
\end{table*}%

\subsection{Baselines}
We select the following methods as baselines and divide them into three categories: SLM methods, LLM models and SLM+LLM methods.
\noindent{\textbf{For the SLM Methods.}}
\noindent \textbf{RoBERTa}~\cite{liu2019roberta} is an extension of BERT~\cite{devlin2018bert} which employs dynamic masking strategies and larger batch sizes during pre-training for natural language understanding.
\textbf{EANN}~\cite{wang2018eann} firstly utilizes a discriminator to derive event-invariant features for multi-domain fake news detection.
\textbf{M$^{3}$FEND}~\cite{zhu2022memory} proposes a memory-guided multi-view framework to address the problem of domain shift and domain labeling incompleteness.
\textbf{FTT}~\cite{hu2023learn} adapts the model to future data by forecasting the temporal distribution patterns of news data.

\noindent{\textbf{For the LLMs.}}
\noindent \textbf{Llama2}~\cite{touvron2023llama} is an auto-regressive, decoder-only large language model based on the Transformer architecture.
\textbf{Llama3} uses the group query attention and a tokenizer with 128K words,  based on Llama2.
\textbf{GPT3.5} is a large language model, which
has shown a surprising ability to solve NLP tasks without finetuning. We conduct zero-shot and few-shot experiments on these LLMs. In the few-shot experiments, we employed a 2-shot setup, wherein each LLM is provided with two real news samples and two fake news samples from test set as demonstrations.

\noindent{\textbf{LLM+SLM Methods.}}
ARG~\cite{hu2024bad} designs an adaptive rationale guidance network for fake news detection, in which SLMs selectively acquire information from LLM's rationales to enhance detection ability.

\subsection{Implementation Details}
Since we focus more on demonstrating the effectiveness of our framework MRCD, we use a simple pre-trained RoBERTa model~\cite{liu2019roberta} and an advanced FTT~\cite{hu2023learn} model as the SLM. To verify the capabilities of different LLMs and their impact on MRCD, we select Llama2-7B, Llama3-8B and GPT-3.5 as the LLMs. To make a fair comparison, we replace the text feature extractors in the baseline SLMs architectures with a pre-trained RoBERTa. We set confidence threshold $\omega$ to {0.8}, batch size to 32, round threshold $\mathcal{N}$ to 3, number of demonstrations $k$ to {4}. Our proposal is trained on 4 NVIDIA 3090 GPUs. The AdamW with a weight decay of 1e-4 is used as the optimizer and the initial learning rate is set to 1e-3.

\begin{table*}[t]
  \centering

  \vspace{-2mm}
  \renewcommand{\arraystretch}{0.75}
    \begin{tabular}{l|rrrr|rrrr}
    \toprule
    \multicolumn{1}{c|}{\multirow{2}[4]{*}{Variant Models}} & \multicolumn{4}{c|}{Pheme}    & \multicolumn{4}{c}{Twitter16}   \\
\cmidrule{2-9}          & \multicolumn{1}{c}{Acc} & \multicolumn{1}{c}{Pre} & \multicolumn{1}{c}{Rec} & \multicolumn{1}{c|}{F1} & \multicolumn{1}{c}{Acc} & \multicolumn{1}{c}{Pre} & \multicolumn{1}{c}{Rec} & \multicolumn{1}{c}{F1} \\
    \midrule
    MRCD(Llama3-Based) &0.788       &0.700       & 0.900      & 0.786      & 0.772      & 0.765      &0.775       & 0.770          \\
    \midrule
    w/o Demonstration & 0.762      & 0.691      & 0.920      &0.774      & 0.680      & 0.808      & 0.675      &0.735       \\
    \textit{w/o Search Engine} &0.774       &0.712       & 0.850      &0.771        &0.763       & 0.754      & 0.761      & 0.757  \\
    \textit{w/o News Corpus}&0.772       & 0. 691     &0.883       &0.778 &0.758       &  0.765     & 0.754      &0.759    \\
    \midrule
    w/o Knowledge & 0.769      &  0.695     & 0.831      &  0.752     &0.761       &0.762       &0.734       &0.748        \\
    w/o Multi-Round &0.717       &0.702      &  0.789     & 0.741      & 0.678      & 0.662      &0.704       &0.682         \\
    \bottomrule
    \end{tabular}%
    \caption{Ablation Study on Pheme and Twitter16.}
  \label{tab:ablation_new}%
\end{table*}%

\begin{figure*}[t]
\centering
\includegraphics[scale=0.82]{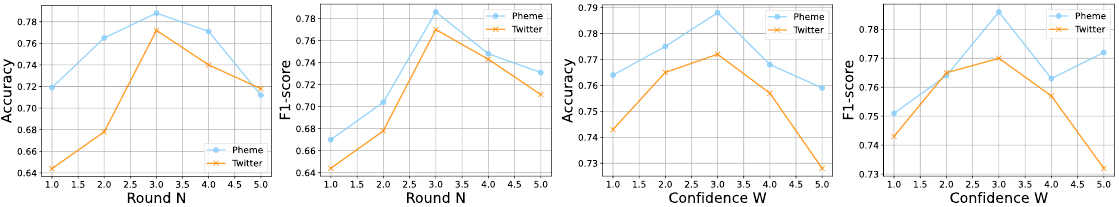}
    \setlength{\abovecaptionskip}{1mm}
    \setlength{\belowcaptionskip}{-2mm}
    \caption{Hyper-parameter sensitivity analysis of $\mathcal{N}$ and $\omega$.}
    \label{fig:sensitive}
\end{figure*}
\subsection{Experimental Results}
Table~\ref{main_new} illustrates the comparison between our approach MRCD and the baseline methods across two datasets.
The three LLMs do not achieve satisfactory results in both zero-shot and few-shot settings, indicating that directly using LLMs for emergent fake news detection is not advisable. However, the application of all three LLMs contributes significantly to the success of MRCD. The Llama3+FTT model achieves the best results, demonstrating that the intrinsic capabilities of the LLM are crucial within our framework.
MRCD(Llama3+RoBERTa) achieves an increase in accuracy of 7.4\% and 12.8\%, and an increase in f1-score of 5.2\% and 12.5\% compared to directly applying the SLM RoBERTa on Pheme and Twitter16 respectively. This demonstrates that MRCD significantly enhances the ability of SLMs to detect emergent news events by effectively utilizing the collaborative paradigm between LLMs and SLMs. 
Furthermore, despite using only a standard RoBERTa model as our SLM rather than a specialized fake news detection model, MRCD still outperforms existing SOTA SLMs for emergent fake news detection and ARG, which fully demonstrates the effectiveness of our approach.  
Notably, the chosen SLM RoBERTa in our framework can be replaced with other more advanced models specifically designed for emergent fake news detection. For instance, MRCD(Llama3+FTT) also proves the detection accuracy significantly compared to FTT, further demonstrating the universality of our framework as it can accommodate any more advanced SLM and LLM to improve the detection ability.

\subsection{Multi-round Learning Analysis}
To verify the effectiveness of multi-round learning, we conduct experiments on the inference results of both the LLM and SLM after each round of learning. Additionally, we record the changes in the number of samples in the clean data pool $D_{clean}$ and noisy data pool $D_{noisy}$ after each round. Here we use Llama3-8B as the LLM, RoBERTa as the SLM and conduct experiments on Twitter16 to investigate the changes. The results are shown in Table~\ref{tab:multi_round}.

In the first round, after undergoing knowledge retrieval from external knowledge bases and demonstration retrieval from online engines and news corpus, the LLM shows significant improvement compared to its zero-shot inference results. This demonstrates that our two-stage retrieval effectively enhances the LLM's detection capabilities, achieving performance close to that of few-shot settings even without labeled samples. In the second and third round, benefiting from the clean samples used as demonstrations for the LLM and finetuning the SLM, both models show significant improvement in detection capabilities compared to the first round. Additionally, data from the noisy pool $D_{noisy}$ gradually transitioned to the clean pool $D_{clean}$. This demonstrates that our multi-round learning approach effectively leverages pseudo-labeled samples to enhance the generalization capabilities of both LLM and SLM for emergent news events, thereby improving the detection accuracy and reliability of the pseudo labels. However, starting from the fourth round, the performance of the SLM experiences a noticeable decline. This is due to an increased number of noisy samples $D_{noisy}$ contaminating the clean sample pool $D_{clean}$, which are then used for finetuning the SLM. In contrast, the LLM is less affected because in-context learning focuses more on the content of the demonstrations rather than the accuracy of the labels, which further supports the feasibility of extracting content from external sources and using pseudo-labels as demonstrations. 
More analysis is in Appendix A.3.

\subsection{Ablation Study}
To investigate the role of each module in MRCD, we conduct ablation experiments for five modules, as shown in Table~\ref{tab:ablation_new}. "w/o Demonstration" denotes the implementation without retrieving demonstrations from online search engines or news corpus for in-context learning. We further conduct experiments with "w/o Search Engine" and "w/o News Corpus" separately to further validate the roles of these two demonstration sources.
"w/o Knowledge" represents the model results after removing the knowledge retrieval module. "w/o Multi-Round" denotes the model not utilizing multi-round learning but using the clean data $D_{clean}$ from the first round to finetune the SLM and make the prediction.

When we remove the retrieval of demonstrations from the external sources, MRCD exhibits a significant decrease in performance across two datasets, especially on the Twitter dataset. This underscores the importance of providing demonstrations for the LLM through in-context learning. Furthermore, to further validate the impact of search engines and news corpus, we conduct ablation experiments on each source separately. The results indicate that eliminating either source led to d decline in model performance, highlighting the necessity of leveraging diverse external sources as demonstrations providers. To validate the impact of knowledge retrieval, we conduct an ablation study by removing the knowledge retrieval component. We observe that the model's performance declined across the two datasets, indicating that knowledge retrieval is crucial for enhancing the model's effectiveness. Finally, we also evaluate the performance without employing multi-round learning. We find that in the absence of multi-round learning, the model struggles to effectively utilize the pseudo-labeled data to adapt to emergent news events, resulting in suboptimal performance.

\subsection{Parameter Sensitivity Analysis}
Here we conduct hyper-parameter sensitivity analysis on the weights of two parameters:  the threshold $\mathcal{N}$ of multi-round learning and the confidence thresholds $\omega$ of data selection. We observe that when the threshold $\mathcal{N}$ of multi-round learning is under 3, MRCD does not perform well in detection because it cannot extract enough clean samples in $D_{clean}$ to fine-tune the SLM and provide demonstrations for the LLM's in-context learning. Conversely, when the threshold $\mathcal{N}$ is larger than 3, too many noisy samples may be incorporated into the clean pool $D_{clean}$, leading to a decline in the SLM's judgment capability. The same applies to confidence threshold $\omega$: when $\omega$ is less than 0.8, the clean pool $D_{clean}$ contains too much noisy samples, reducing the SLM's performance. On the other hand, when $\omega$ is larger than 0.8, there are too few samples to fine-tune the SLM and provide demonstrations for the LLM, resulting in poor accuracy as both models cannot adequately adapt to new news events.

\section{Conclusion}
\label{Conclusion}
Emergent fake news detection primarily faces the challenges of inconsistent distribution of emerging data and the lack of annotation. To address these challenges,
in this paper we propose a multi-round collaboration framework between LLM and SLM for emergent fake news detection, dubbed MRCD. We propose a two-stage retrieval module, a data selection module, and a multi-round collaboration module to enhance detection capability in unsupervised emergent news events settings.
Extensive experiments on two real-world datasets have proven the effectiveness of our model MRCD.





\newpage
\appendix
\section{{Appendix}}
\section{Experimental Details}
\subsection{Prompt Setting}
We provide our prompt design on the Pheme and Twitter dataset
for the initial demonstration generation step and the query step.
\newline
\textit{"system":You are a useful assistant, working as a fake news detector on the dataset. This news dataset consists of a mix of fake news reviews and real news reviews. Your task is to make a binary detection, categorizing the news as real or fake based on your judgment of it. The category is divided into two types: real or fake.}
\newline
\textit{"role":Give a news:\{demonstration\}.Do you think the above news is more real or more fake?.}
\newline
\textit{"assistant":random(label)}
\newline
\textit{"role":Give you some extra information from wikidata: \{wikidata\}.Then, give you a news to detect: \{news\}.Do you think the above news is more real or more fake?please answer in a single line with real or fake.}
\begin{table}[htbp]
  \centering
  \caption{Basic information of two datasets.}
    \begin{tabular}{l|rrr|rrr}
    \toprule
    \multicolumn{1}{c|}{\multirow{2}[4]{*}{Dataset}} & \multicolumn{3}{c|}{Pheme} & \multicolumn{3}{c}{Twitter16} \\
\cmidrule{2-7}          & \multicolumn{1}{l}{Train} & \multicolumn{1}{l}{Val} & \multicolumn{1}{l|}{Test} & \multicolumn{1}{l}{Train} & \multicolumn{1}{l}{Val} & \multicolumn{1}{l}{Test}  \\
    \midrule
    Real  & 2895      & 704      & 231      & 4046      & 1025      & 719       \\
    Fake  & 1371      & 363      & 238      & 4793      & 1185      & 707      \\
    Total & 4266      & 1067      & 469      & 8839      & 2210      & 1426       \\
    \bottomrule
    \end{tabular}%
  \label{tab:dataset_v2}%
\end{table}%
         
\begin{table}[htbp]
  \centering
  \caption{A comparison between clean data pool and noisy data pool.}
  \resizebox{\linewidth}{!}{
    \begin{tabular}{c|c|rrrr|rrrr}
    \toprule
    \multirow{2}[4]{*}{Round} & \multirow{2}[4]{*}{Model} & \multicolumn{4}{c|}{Clean Data Pool} & \multicolumn{4}{c}{Noisy Data Pool} \\
\cmidrule{3-10}          &       & \multicolumn{1}{c}{Acc} & \multicolumn{1}{c}{Pre} & \multicolumn{1}{c}{Rec} & \multicolumn{1}{c|}{F1} & \multicolumn{1}{c}{Accuracy} & \multicolumn{1}{c}{Pre} & \multicolumn{1}{c}{Rec} & \multicolumn{1}{c}{F1} \\
    \midrule
    \multirow{2}[2]{*}{1} & SLM   &0.758      &0.735       &0.840       &0.786       &0.538      &0.531       &0.570       &0.562  \\
          & LLM   &0.758      &0.735       &0.840       &0.786       &0.316       &0.251       &0.256       &0.268  \\
    \midrule
    \multirow{2}[2]{*}{2} & SLM   &0.870      &0.863       &0.952       &0.905      &0.599      &0.553       &0.582       &0.598  \\
          & LLM   &0.870      &0.863       &0.952       &0.905       &0.314       &0.281       & 0.274      &0.291  \\
    \midrule
    \multirow{2}[2]{*}{3} & SLM   &0.879       &0.868       &0.949       & 0.907       & 0.654      &0.548       &0.864       &0.671  \\
          & LLM   &0.879       &0.868       &0.949       & 0.907      & 0.264      & 0.241      &0.236       & 0.245 \\
    \midrule
    \multirow{2}[2]{*}{4} & SLM   &0.758       &0.735       &0.845       & 0.786       &0.636       &0.582       &0.724       &0.635  \\
          & LLM  &0.758       &0.735       &0.845       & 0.786       & 0.294      & 0.263      &0.224       & 0.258  \\
    \midrule
    \multirow{2}[2]{*}{5} & SLM   &0.742       &0.753       &0.773       &0.764       & 0.642      &0.576       & 0.703      &0.643  \\
          & LLM   &0.742       &0.753       &0.773       &0.764       &0.302       &0.271       &0.228       &0.261  \\
    \bottomrule
    \end{tabular}%
    }
  \label{tab:clean_noisy}%
\end{table}%

\subsection{Dataset}
\label{A.2}
Here we provide detailed information about the datasets in table~\ref{tab:dataset_v2}.
\begin{table}[htbp]
  \centering
  \caption{Basic information of two datasets.}
    \begin{tabular}{l|rrr|rrr}
    \toprule
    \multicolumn{1}{c|}{\multirow{2}[4]{*}{Dataset}} & \multicolumn{3}{c|}{Pheme} & \multicolumn{3}{c}{Twitter16} \\
\cmidrule{2-7}          & \multicolumn{1}{l}{Train} & \multicolumn{1}{l}{Val} & \multicolumn{1}{l|}{Test} & \multicolumn{1}{l}{Train} & \multicolumn{1}{l}{Val} & \multicolumn{1}{l}{Test}  \\
    \midrule
    Real  & 2895      & 704      & 231      & 4046      & 1025      & 719       \\
    Fake  & 1371      & 363      & 238      & 4793      & 1185      & 707      \\
    Total & 4266      & 1067      & 469      & 8839      & 2210      & 1426       \\
    \bottomrule
    \end{tabular}%
  \label{tab:dataset_v2}%
\end{table}%

\subsection{Comparison Between Clean and Noisy Data}
\label{A.3}
To further analyze the clean sample pool $D_{clean}$ and noisy sample pool $D_{noisy}$, we conduct an analysis of accuracy and other metrics in each round as shown in Table~\ref{tab:clean_noisy}. From the first round, the accuracy and other metrics in the clean pool $D_{clean}$ are significantly higher than those in the noise pool $D_{noisy}$, proving the reliability of our data selection module. Meanwhile, as the number of rounds increased, the accuracy in the clean pool initially improved significantly. However, when too many rounds are conducted, the performance of the model began to decline due to the inclusion of noisy data in the clean pool. This demonstrates the importance of round threshold $\mathcal{N}$ and confidence threshold $\omega$.
\begin{table}[htbp]
  \centering
  \caption{Analysis of pseudo labels for demonstrations.}
  \resizebox{\linewidth}{!}{
    \begin{tabular}{l|rrrr}
    \toprule
    Pseudo labels & \multicolumn{1}{l}{Accuracy} & \multicolumn{1}{l}{Precision} & \multicolumn{1}{l}{Recall} & \multicolumn{1}{l}{F1-Score} \\
    \midrule
    LLM (zero-shot) &0.562       &0.574       & 0.531      &0.552  \\
    LLM (few-shot) &0.622       & 0.607      &0.717       &0.658  \\
    \midrule
    Reliable/Unreliable &0.604       &0.588       & 0.722      &0.648  \\
    Convincing/Incredible & 0.608      & 0.586      &0.714       &0.644  \\
    \midrule
    True/False &0.598       &0.595       & 0.710      &0.647  \\
    Real/Fake &0.589       &0.564       & 0.795      & 0.642 \\
    \midrule
    No-label &0.578       &0.579       & 0.692      &0.634  \\
    \midrule
    Training Dataset &0.584       &0.592      & 0.645      &0.626  \\
    \bottomrule
    \end{tabular}%
    }
  \label{tab:pseudo}%
\end{table}%

\subsection{Pseudo Labels for Demonstrations.}
\label{Pseudo Appendix}
In the main experiment, we use synonymous replacements \textit{realistic/unrealistic} as pseudo-labels for demonstrations to conduct in-context learning for the large model. In this section, we attempt to experiment with different pseudo-labels to explore their impact on the zero-shot results of the LLM. The results are shown in Table~\ref{tab:pseudo}.  For pseudo labels, we use \textit{Reliable/Unreliable}, \textit{Convincing/Incredible} as synonyms pseudo labels. We use \textit{Real/Fake} and \textit{True/False} as the same label with the output of fake news detection. In addition, we conduct experiments with no-labeled news and retrieve demonstrations from training dataset.

Experimental data indicates that providing a few examples to LLMs significantly enhances their performance compared to zero-shot conditions. Utilizing synonyms as pseudo labels achieves better results than using direct task labels, due to the avoidance of copy effects in large models. Without pseudo labels on the demonstrations, performance degrades somewhat, as the absence of labels fails to fully activate the large language model's capability for detecting fake news. Selecting labeled training datasets as demonstrations does not match the effectiveness of pseudo labels, highlighting that the relevance of demonstration content to the test content is more crucial than the authenticity of the labels themselves. This underscores the importance of MRCD retrieving the most relevant examples from online search engines and news corpus.

\begin{figure*}[t]
\centering
\includegraphics[scale=0.66]{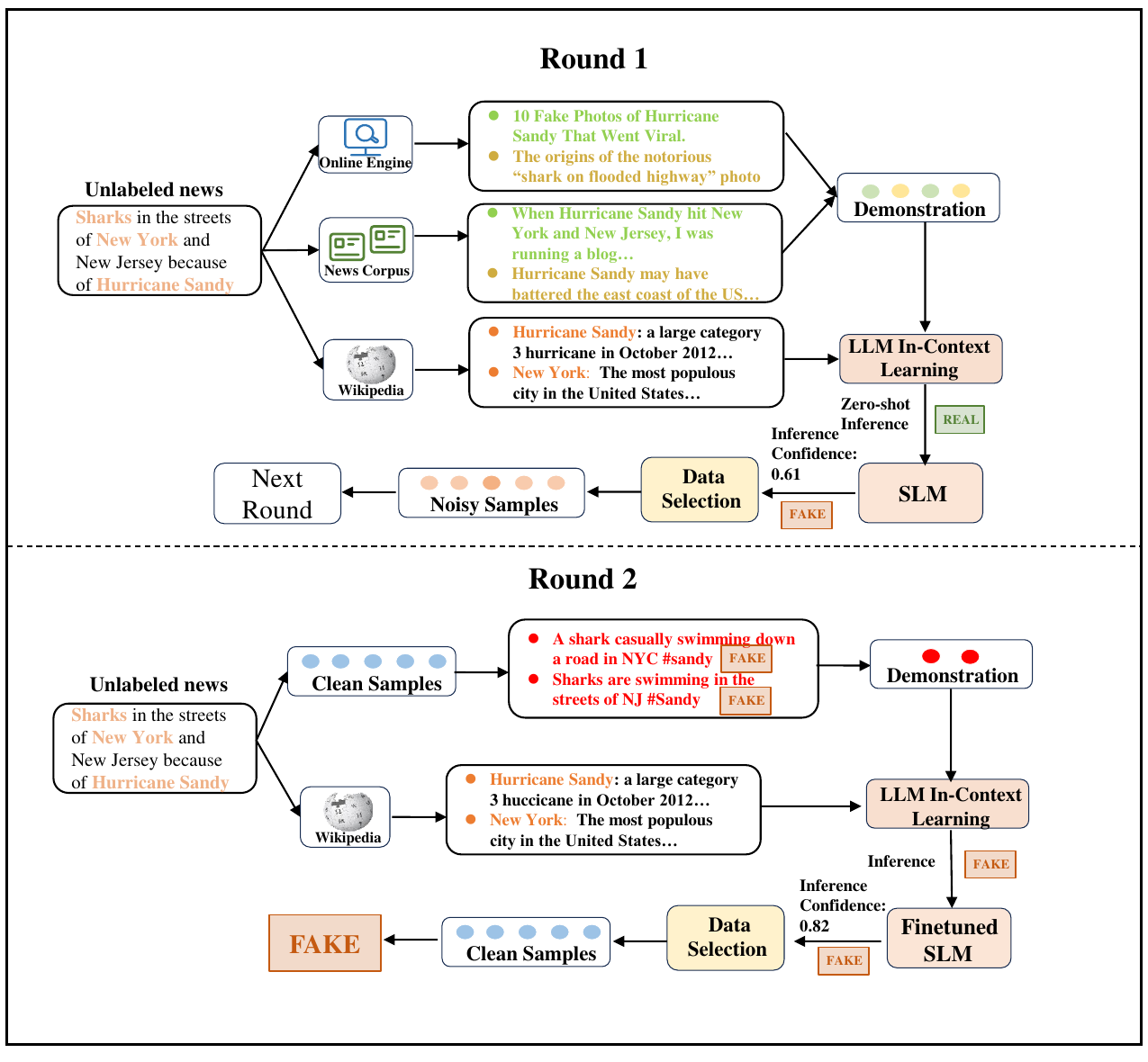}
    \caption{A case study of how MRCD works.}
    \label{case_study}
\end{figure*}

\subsection{Notations}
We make a detailed description of all notations in our paper for the clarity and readability of MRCD at Table~\ref{notations}.
\begin{table}[htbp]
    \centering
    \caption{The descriptions of annotations.}
    \label{notations}
    \begin{tabular}{|l|l|}
    \hline
        notations & description \\ \hline
        $\epsilon$ & A set of news events \\ \hline
        $e_i$ & The $i_{th}$ event \\ \hline
        $t_i$ & The timestamp of $e_i$ \\ \hline
        ${X_e^s,Y_e^s}$ & Labeled past events \\ \hline
        ${X_e^t}$ & Unlabeled emergent events \\ \hline
        $x$ & News under detection \\ \hline
        $\mathcal{W}$ & Online search engine \\ \hline
        $\mathcal{C}$ & News corpus \\ \hline
        $N_w$ & Retrieved news from search engines $\mathcal{W}$ \\ \hline
        $N_c$ & Retrieved news from news corpus $\mathcal{C}$ \\ \hline
        $N_k$ & Top-K retrieved news \\ \hline
        $\mathcal{D}$ & Demonstrations \\ \hline
        $\mathcal{K}$ & Retrieved knowledge from Wikipedia \\ \hline
        $D_{clean}$ & A set of labeled clean data \\ \hline
        $D_{noisy}$ & A set of unlabeled noisy data \\ \hline
        $\mathcal{N}$ & The threshold of round \\ \hline
    \end{tabular}
\end{table}

\subsection{Case Study}
In order to help understand the workflow of our framework MRCD, we provide a case study to detailedly introduce how our framework MRCD helps the SLMs to successfully detect the authenticity of news, which can be seen at Figure~\ref{case_study}.

In the first round, the online retrieval and news corpus retrieval is conducted for the unlabeled news to select most relevant data with pseudo labels as demonstrations for in-context learning. A LLM is utilized to extract key entities from news and retrieve knowledge information from wikipedia to help LLM understand the background of news. After in-context learning, LLM ans pretrained SLM will give their judgements. As LLM's perdiction is real and SLM's prediction is fake, the data is fed into noisy samples for another iteration.

In the second round, different from first round utilizing online search engine and news corpus to retrieve demonstrations, MRCD directly utilizes the clean data pool to select most relevant news with their predicted labels as demonstrations for LLM's in-context learning. After the second round in-context learning, LLM's prediction aligns with SLM's prediction while the SLM's confidence is 0.82 larger than the 
confidence threshold $\omega$, the news is labeled as fake and fed into clean samples.

\subsection{Limitation}
Despite the promising performance, we note there are certain \textit{limitations} to MRCD. Firstly, MRCD requires a two-stage retrieval module to retrieve demonstrations and external knowledge, which may consume time and resources. Secondly, the selection of hyper-parameters in MRCD requires manual selection rather than automatic selection. This remains a challenge for future work.
\section*{Acknowledgments}
This research is supported by the National Natural Science
Foundation of China (No.62272025 and No.U22B2021), and
the Fund of the State Key Laboratory of Software Development Environment.

\bibliography{aaai25}

\end{document}